# A Vision-and-Knowledge Enhanced Large Language Model for Generalizable Pedestrian Crossing Behavior Inference

Qingwen Pu[a], Kun Xie[a],*, Hong Yang[b], and Guocong Zhai[c]

[a] Transportation Informatics Lab, Department of Civil and Environmental Engineering, Old Dominion University, Norfolk, VA 23529, United States
[b] Department of Electrical and Computer Engineering, Old Dominion University, Norfolk, VA, United States
[c] School of Transportation and Logistics, National Engineering Laboratory of Integrated Transportation Big Data Application Technology, National and Local Joint Engineering Research Center of Integrated Transportation Intelligence, Southwest Jiaotong University, Chengdu 611756, China
* Corresponding Author; Email: kxie@odu.edu

## Abstract

Existing paradigms for inferring pedestrian crossing behavior, ranging from statistical models to supervised learning methods, demonstrate limited generalizability and perform inadequately on new sites. Recent advances in Large Language Models (LLMs) offer a shift from numerical pattern fitting to semantic, context-aware behavioral reasoning, yet existing LLM applications lack domain-specific adaptation and visual context. This study introduces Pedestrian Crossing LLM (PedX-LLM), a vision-and-knowledge enhanced framework designed to transform pedestrian crossing inference from site-specific pattern recognition to generalizable behavioral reasoning. By integrating LLaVA-extracted visual features with textual data and transportation domain knowledge, PedX-LLM fine-tunes a LLaMA-2-7B foundation model via Low-Rank Adaptation (LoRA) to infer crossing decisions. PedX-LLM achieves 82.0% balanced accuracy, outperforming the best statistical model (Hierarchical Logistic Regression) by 7.9 percentage points and the best supervised learning method (CatBoost) by 3.0 percentage points. Results demonstrate that the vision-augmented module contributes a 2.9% performance gain by capturing the built environment and integrating domain knowledge yields an additional 4.1% improvement. To evaluate generalizability across unseen environments, cross-site validation was conducted using site-based partitioning. The zero-shot PedX-LLM configuration achieves 66.9% balanced accuracy on five unseen test sites, outperforming the baseline data-driven methods by at least 18 percentage points. Incorporating just five validation examples via few-shot learning to PedX-LLM further elevates the balanced accuracy to 72.2%. PedX-LLM demonstrates strong generalizability to unseen scenarios, confirming that vision-and-knowledge-enhanced reasoning enables the model to mimic human-like decision logic and overcome the limitations of purely data-driven methods.

## 1. Introduction

Pedestrian safety remains a critical public health challenge in the United States. In Virginia, for instance, annual pedestrian fatalities increased by 41.3% and serious injuries by 32.2% from 2016 to 2023, reflecting a troubling upward trend. According to a Virginia Transportation Research Council report (FHWA/VTRC 25-R6), approximately 70% of pedestrian fatalities in Virginia occur at mid-block locations, where pedestrians frequently choose to cross outside designated crosswalks (Xie et al. 2024). Pedestrian crossing inference models provide a systematic framework for understanding the factors that most strongly influence crossing location selection and for inferring the scenarios in which risky crossing behaviors are likely to occur (Papadimitriou *et al.* 2016). Such models can inform targeted interventions and guide infrastructure design improvements aimed at reducing mid-block fatalities.

Existing paradigms for inferring pedestrian crossing behavior, ranging from statistical models to supervised learning methods, demonstrate limited generalizability and perform inadequately on new sites (Ham *et al.* 2024). Most statistical models impose linear assumptions that limit their ability to capture complex interactions among safety factors (Kyriazos and Poga 2024). For example, these models struggle to represent how roadway width effects vary with pedestrian age or how lighting influence depends on traffic conditions. These parametric assumptions often fail to represent the nonlinear, context-dependent heuristics that characterize real-world pedestrian decisions. Supervised learning methods treat behavioral inference as numerical pattern recognition task, learning statistical associations between input features and crossing outcomes from training data (Pu *et al.* 2025c). However, these approaches focus on fitting site-specific patterns with little attention to underlying reasoning processes, which limits their ability to generalize beyond their training environments (Elalouf *et al.* 2023). As a result, these approaches primarily learn site-specific patterns rather than universal behavioral principles, which limits their applicability to unseen contexts.

Large Language Models (LLMs) enable a paradigm shift from numerical pattern fitting to semantic behavioral reasoning (Fahad *et al.* 2025). Trained on vast corpora of human-generated text, LLMs encode implicit knowledge of human cognition, reasoning patterns, and social norms (Yao *et al.* 2024). This foundation allows LLMs to approximate human-like reasoning by capturing context-sensitive, non-linear decision processes that conventional data-driven methods struggle to formalize (Huang *et al.* 2025). Unlike conventional methods that require site-specific calibration, LLMs can adapt to new contexts through few-shot learning or targeted prompting (Hang *et al.* 2025). By leveraging generalized behavioral priors, LLMs infer pedestrians' decisions in unseen environments with minimal additional data, offering a generalizable and data-efficient alternative to conventional methods (Kim *et al.* 2025).

However, applying LLMs for pedestrian behavior inferences faces fundamental challenges. Generic pre-trained models lack the transportation-specific knowledge necessary for distinguishing true contributing factors from spurious correlations (Liu *et al.* 2025). Existing LLM applications often fail to jointly reason over visual and textual information, limiting their ability to capture how built environment features—such as road geometry and traffic conditions—interact with individual characteristics to shape pedestrian behavior (Blečić *et al.* 2024). Additionally, using commercial LLM APIs (e.g., GPT-5, Gemini, Claude) raises privacy concerns as sensitive visual and behavioral data must be transmitted to external servers (Le *et al.* 2025). These limitations highlight the need for a privacy-preserving framework that integrates visual reasoning with domain knowledge to support accurate and context-aware pedestrian crossing inference.

This study introduces PedX-LLM, a vision-and-knowledge enhanced LLM that transitions pedestrian crossing prediction from numerical pattern recognition to generalizable behavioral reasoning, as shown in Figure 1. The framework integrates field observations and a LLaVA-based vision module with domain-knowledge prompts for grounding behavioral reasoning and employs Shapley-based attribution to decode the decision-making process explicitly. PedX-LLM utilizes Low-Rank Adaptation (LoRA) to fine-tune the LLaMA-2-7B model, enabling fully local training and deployment to ensure strict data privacy. The main contributions of this study include:

- Combine satellite imagery with textual data to capture the combined effects of the built environment and individual characteristics on crossing decisions.
- Integrate domain knowledge into behavioral reasoning via parameter-efficient fine-tuning, which transforms opaque pattern recognition into interpretable inference.
- Demonstrate strong generalizability to unseen scenarios, showing that a vision-and-knowledge enhanced reasoning mitigates distribution shifts and supports reliable inference across heterogeneous contexts.

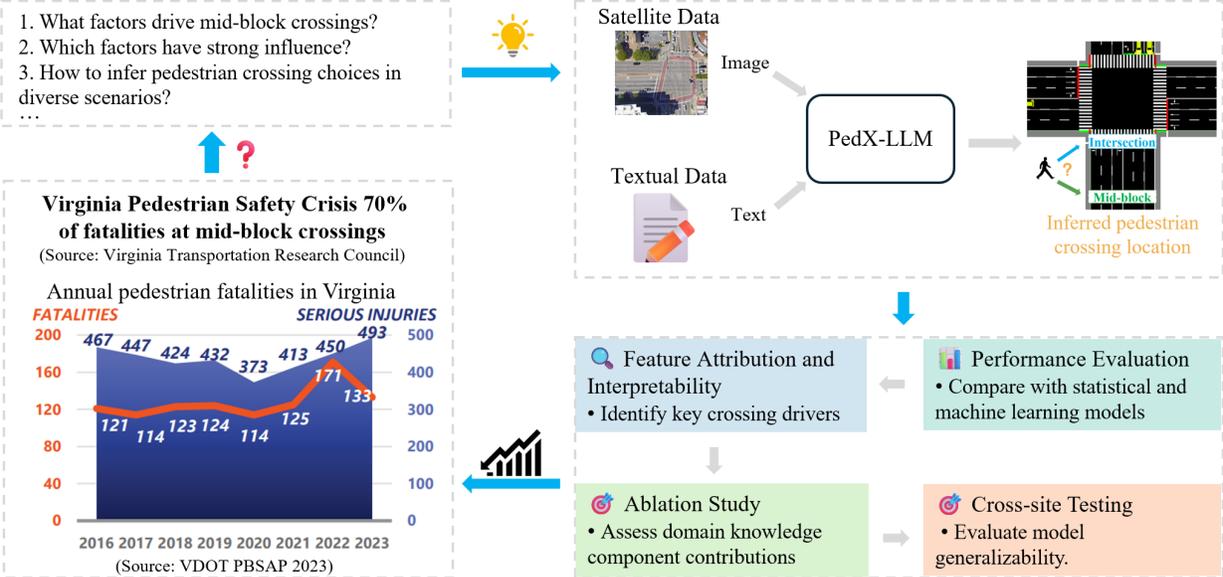

**Figure 1: Overview of the conceptual framework**

## 2. Literature Review

### 2.1. Pedestrian Crossing Behavior Inference

The decision to cross mid-block rather than at a designated intersection represents a fundamental trade-off between convenience and safety. This choice emerges from complex interactions between individual characteristics (such as age and gender) and environmental contexts (such as lighting and traffic control) (Holland and Hill 2010, Osei *et al.* 2024). However, behavioral variability across urban environments complicates risk assessment because factors driving mid-block crossing at one site often differ at another site (Zhu *et al.* 2025). Consequently, inference frameworks that derive behavioral mechanisms from observational data are essential for guiding data-driven policy decisions, including infrastructure improvements and targeted behavioral interventions (Singh *et al.* 2024).

Traditional statistical frameworks provide the foundational approach for quantifying determinants of pedestrian crossing location choices. Discrete choice methods, particularly binary and multinomial logistic regression, are widely utilized to estimate the probability of mid-block crossing based on observable variables (Ahsan *et al.* 2025). The main advantage of these parametric approaches is interpretability, as the resulting coefficients provide direct estimates of marginal effects, enable hypothesis testing, and support established behavioral theories (Gu *et al.* 2024). However, these models rely on linear assumptions that constrain their ability to infer complex nonlinear interactions inherent in pedestrian-environment dynamics (Hagenaars *et al.* 2024). Furthermore, statistical approaches struggle with high-dimensional feature spaces that include diverse demographic and environmental variables (Koldasbayeva *et al.* 2024). Supervised learning approaches, including machine learning algorithms and deep learning architectures, offer a data-driven alternative to overcome these statistical modeling constraints.

Machine learning algorithms overcome the parametric limitations of statistical frameworks by inferring nonlinear relationships and complex feature interactions in pedestrian crossing behavior (Pu *et al.* 2025b). Random Forest classifiers utilize feature importance rankings to identify influential predictors (Yuan *et al.* 2023). However, this method does not infer underlying behavioral mechanisms or establish causal links between demographics and risk perception (Jacobucci *et al.* 2023). Gradient boosting frameworks, including XGBoost and CatBoost, effectively handle non-linear interactions and categorical data (Zhang and Jánošík 2024). However, these black-box models predict outcomes without explaining why specific groups, such as older pedestrians, prefer intersections (Rafie *et al.* 2025). Furthermore, Support Vector Machines generate complex decision boundaries that cannot effectively decouple individual-level risk tolerance from built environment effects (Tanveer *et al.* 2024). Consequently, these limitations necessitate the adoption of deep learning architectures capable of capturing higher-order feature interactions.

Deep learning architectures offer advanced capabilities for capturing hierarchical patterns in pedestrian behavior (Huang *et al.* 2024). Multi-layer Perceptrons (MLP) utilize fully connected networks to process tabular demographic and site data. However, without regularization or architectural constraints, these models frequently overfit to specific training environments (Przybyła-Kasperek and Marfo 2024). For instance, models with fixed parameters optimized for specific arterial configurations often exhibit performance degradation when applied to environments with distinct geometric features (Pu *et al.* 2025a). This failure indicates that the model memorizes local geometric features rather than inferring generalizable behavioral principles (Shanthini *et al.* 2024). TabNet employs sequential attention mechanisms to learn feature importance from structured data (Li 2025). While this approach reduces manual feature engineering, it requires large datasets to converge. Additionally, it processes textual descriptions separately from visual spatial contexts, resulting in incomplete environmental reasoning (Jiang *et al.* 2025). Consequently, these deep learning models suffer from poor cross-site generalization and limited interpretability despite the use of attention mechanisms.

Despite the evolution from statistical inference to deep learning, the field remains constrained by the inherent trade-off between interpretability and predictive accuracy. Moreover, existing approaches exhibit limited capacity to infer behavioral mechanisms that generalize across different urban environments (Geiger *et al.* 2025). These deficiencies underscore the need for a paradigm shift toward models that enable generalizable inference of causal mechanisms across different urban environments.

## 2.2. Large Language Models for Pedestrian Behavior

The emergence of LLMs offers a significant opportunity to advance pedestrian safety research through their advanced ability to process and reason over complex multimodal information (Karim *et al.* 2025). These models synthesize vast quantities of unstructured data, such as demographic attributes and narrative descriptions of crossing circumstances, and align them with structured roadway characteristics to generate holistic behavioral insights (Li *et al.* 2025). In the context of pedestrian behavior analysis, LLMs facilitate three critical functions: extracting pedestrian-vehicle conflict patterns from textual crash narratives, recognizing behavioral strategies within crossing descriptions, and identifying latent risk factors from injury databases (Yan *et al.* 2025). This capacity to integrate heterogeneous data sources enables more holistic understanding of pedestrian crossing decision mechanisms than traditional methods (Pu *et al.* 2026).

Current LLM research in pedestrian safety predominantly employs generic prompt engineering (Abbasi and Rahmani 2025). Text-based models learn crossing patterns from narrative descriptions without incorporating prior empirical knowledge (Strömel *et al.* 2024). Consequently, these architectures capture statistical correlations but fail to understand the causal mechanisms driving pedestrian crossing decisions (Yang *et al.* 2025). Current methodologies lack the integration of specific behavioral theories. At the individual level, models frequently overlook heterogeneity in age-related risk perception and group dynamics (Deng *et al.* 2025). Similarly, gender-based variations in risk tolerance are often neglected (Mirza *et al.* 2025). Regarding the built environment, physical determinants such as public transit stations and sidewalk width are rarely encoded explicitly (Rodríguez *et al.* 2009). Consequently, the field lacks structured prompt engineering frameworks designed to integrate prior knowledge regarding pedestrian crossing behavior. Moreover, existing LLM studies have not conducted rigorous ablation analyses to quantify the distinct contributions of individual-level factors versus built environment knowledge.

Existing frameworks in pedestrian safety segregate the processing of visual and textual data. Vision models utilize satellite or street view imagery to extract built environment characteristics, such as land use patterns and sidewalk configurations (Zhou *et al.* 2025). However, these visual components operate as independent feature extractors rather than as integrated elements within LLM reasoning logic (Torneiro *et al.* 2025). Simultaneously, textual architectures analyze demographic attributes and behavioral circumstances in isolation from the spatial context (Peykani *et al.* 2025). This independent processing fails to capture the synergistic mechanisms where the built environment interacts with specific pedestrian traits (Yang *et al.* 2024). For instance, older pedestrians prioritize risk avoidance over time savings, often avoiding mid-block crossings on multi-lane arterials (Wilmut and Purcell 2022). However, such synergistic mechanisms are frequently overlooked when visual roadway features and demographic profiles are analyzed in isolation (Sharif *et al.* 2025). Consequently, the field lacks hybrid vision-augmented architectures that can jointly process built-environment features and individual attributes to provide a comprehensive analysis of pedestrian crossing behavior.

Pedestrian behavior analysis requires sensitive personal attributes, including demographic characteristics and geospatial coordinates (Jie *et al.* 2025). However, current studies predominantly utilize commercial cloud-based APIs or generic pre-trained models due to the computational resources required to train proprietary architectures (Raiaan *et al.* 2024, Choi and Chang 2025). This widespread reliance on external services necessitates the transmission of confidential observational data to third-party servers (Ali and Ghanem 2025). Such transmission violates regulatory constraints of transportation agencies, which are often prohibited from sharing granular behavioral datasets owing to liability concerns (Hockstad *et al.* 2025). Consequently, a critical gap exists for frameworks enabling the local training of proprietary, domain-specific LLMs entirely within agency-controlled infrastructure.

### 2.3. Research Gaps

The literature review identifies three critical gaps in pedestrian crossing behavior inference. First, conventional statistical and supervised learning approaches demonstrate limited cross-site transferability (Efron 2020), learning site-specific numerical patterns rather than generalizable behavioral principles. Second, existing LLM applications lack transportation-specific domain adaptation, relying on generic prompting without integrating empirically validated behavioral theories regarding demographic heterogeneity and built environment effects (Liu *et al.* 2025). Third, current frameworks process visual and textual data independently, failing to jointly reason over built environment features and individual attributes within a unified inference architecture (Zhang *et al.* 2025). Additionally, the widespread reliance on commercial cloud-based APIs raises privacy concerns for sensitive behavioral data (Shanmugarasa *et al.* 2025). These gaps underscore the need for a privacy-preserving framework that integrates vision-augmented reasoning with domain knowledge to enable generalizable pedestrian crossing inference across heterogeneous urban environments.

### 3. Data Preparation

Field observations were conducted at 35 mid-block locations that are adjacent to intersections with crosswalks in the Hampton Roads region, Virginia, as shown in Figure 2 (a). This study selected sites using systematic criteria to capture diverse pedestrian crossing contexts where mid-block crossing demand exists but crosswalk infrastructure was absent or distant. Selection criteria included arterials and collectors in urban or suburban areas with notable pedestrian generators such as transit stops, residential areas, educational campuses, parks, healthcare facilities, shopping centers, and parking facilities. All sites maintained a minimum 300-foot distance to the nearest marked crosswalk per Virginia Department of Transportation standard IIM-TE-384.18.

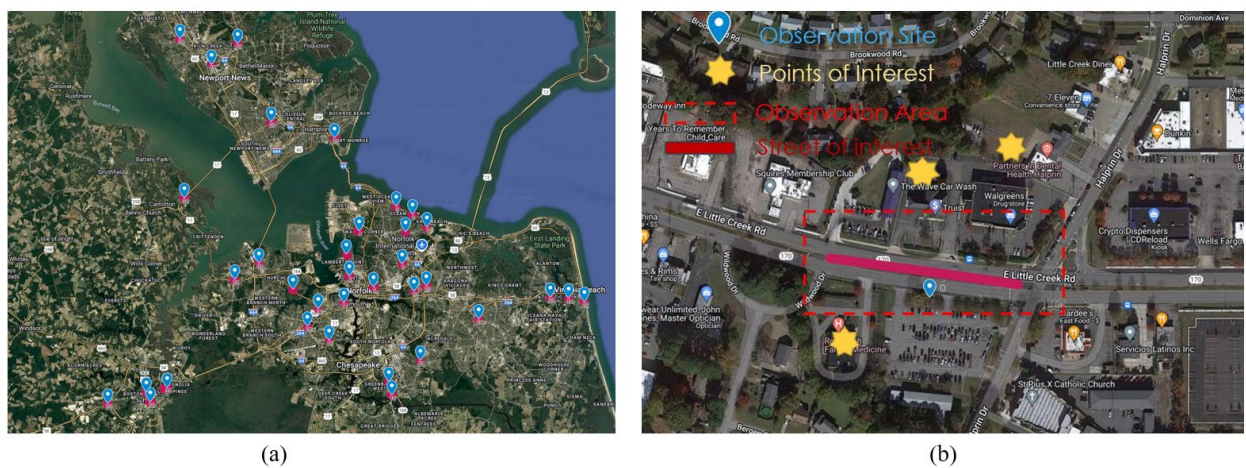

**Figure 2: (a) Distribution of 35 observation sites across Hampton Roads region, Virginia; (b) Layout of one observation site**

The 35 selected sites span multiple jurisdictions within Hampton Roads and represent diverse roadway configurations and land use contexts. Table 1 summarizes the built environment characteristics. These characteristics provide the physical framework within which pedestrian crossing decisions occur and represent key determinants of crossing location choice between intersection and mid-block alternatives.

**Table 1 Built Environment Characteristics at Study Sites (N=35 sites)**

| Variable | Categories/Range | Description |
|---|---|---|
| Number of lanes at mid-block | 2-10 lanes | Mean: 4.5 |
| Speed limit | 25-35mph | 25mph:40%, 30mph:43%, 35mph:17% |
| Presence of raised median | Present/Absent | Present: 37%, Absent: 63% |
| Presence of Public Transit Station | Present/Absent | Present: 77%, Absent: 23% |
| Total width of sidewalk | 10-14ft | Mean: 11.4 ft |
| Land use | 5 categories | Educational-residential (29%), commercial-residential (23%), office-residential (20%), educational-office (14%), and green space-residential (14%) |

Figure 2 (b) depicts a typical observation site configuration, showing the observation site, point of interest, observation area, street of interest. Trained field investigators positioned themselves at predetermined observation points between adjacent intersection crosswalks where potential mid-block crossings could occur. They ensured unobstructed visibility of bidirectional pedestrian movements at both intersection and mid-block locations. Investigators observed each site continuously for six hours to capture temporal variations across daytime and nighttime conditions. Due to seasonal variations in sunset timing, we adapted observation schedules: data collection occurred from 2:00 PM to 8:00 PM between November 24, 2022, and March 31, 2023 (87 daytime hours and 63 nighttime hours), and from 4:00 PM to 10:00 PM during other collection periods (32 daytime hours and 24 nighttime hours). This protocol resulted in approximately 210 total observation hours across 35 sites, covering both daytime and nighttime periods.

For each observed pedestrian, investigators manually documented demographic characteristics (age group and gender), Walking context (walking alone or in a group), crossing behavior (crossing location choice: intersection versus mid-block), weather conditions, lighting conditions at both intersection and mid-block locations, and traffic control features. Traffic control features included signal timing parameters (time interval between consecutive onsets of green time for crossing and green interval for crossing), pedestrian push-button availability and functionality, and left-turn protection phases. Data collection protocols received Institutional Review Board approval (November 3, 2022), ensuring participant anonymity through the exclusion of personally identifiable information. The final dataset comprises 687 pedestrian observations (Table 2). Of these, 259 (37.7%) occurred at mid-block locations, while 428 (62.3%) took place at intersections.

**Table 2 Demographic and Behavioral Characteristics of Observed Pedestrians (N=687)**

| Variable | Category | Frequency | Percentage |
|---|---|---|---|
| **Crossing location** | Intersection (with crosswalks) | 428 | 62.3% |
| | Mid-block (without crosswalks) | 259 | 37.7% |
| Age group | Child (<18) | 104 | 15.1% |
| | Adult (18-65) | 560 | 81.5% |
| | Senior (>65) | 15 | 2.2% |
| | Unsure | 8 | 1.2% |
| Gender | Male | 459 | 66.8% |
| | Female | 210 | 30.6% |
| | Unsure | 18 | 2.6% |
| Walking context | Alone | 435 | 63.3% |
| | In group | 252 | 36.7% |
| Weather condition | Clear/sunny | 325 | 47.3% |
| | Partly cloudy | 207 | 30.1% |
| | Cloudy | 79 | 11.5% |

| | Rain/drizzle | 76 | 11.1% |
|---|---|---|---|
| Lighting at intersection | Present | 583 | 84.9% |
| | Absent | 104 | 15.1% |
| Lighting at mid-block | Present | 630 | 91.7% |
| | Absent | 57 | 8.3% |
| Time interval between consecutive onsets of green time for crossing (s) | 0 s | 60 | 8.7% |
| | 1 to 85 s | 346 | 50.4% |
| | > 85 s | 276 | 40.2% |
| | Missing | 5 | 0.7% |
| Green interval for crossing (s) | 0 s | 60 | 8.7% |
| | 1 to 40 s | 373 | 54.3% |
| | > 40 s | 249 | 36.2% |
| | Missing | 5 | 0.7% |
| Pedestrian push-button available | Yes | 571 | 83.1% |
| | No | 116 | 16.9% |
| Push-button affecting crossing time | Yes | 492 | 86.2% |
| | No | 79 | 13.8% |
| Left turn protection phase | Yes | 471 | 68.6% |
| | No | 216 | 31.4% |

The observational methodology captures authentic crossing behavior unaffected by experimental intervention. It is worth noting that the selected mid-block locations exhibited higher lighting coverage (91.7%) compared to intersections (84.9%), reflecting the continuous arterial lighting prevalent in the study area. However, the study relies on observers' judgment for certain classifications, particularly age estimation. To mitigate classification bias, particularly in age estimation, rigorous investigator training protocols were implemented to ensure the consistent application of criteria.

## 4. Methodology

This study develops the PedX-LLM framework for inferring pedestrian crossing location choices. As illustrated in Figure 3, the architecture integrates multi-modal data collection, structured prompt engineering, and parameter-efficient fine-tuning. Conventional supervised learning approaches rely primarily on numerical feature vectors to predict outcomes without inferring underlying mechanisms. In contrast, our framework leverages LLMs' contextual inference capabilities to integrate heterogeneous data sources, including field observations, satellite imagery, and domain knowledge, enabling the derivation of behavioral mechanisms from multi-modal contexts through a unified text-based pipeline. The framework transforms the binary classification task (intersection crossing versus mid-block crossing) into a natural language generation problem, enabling the model to leverage pre-trained linguistic knowledge while incorporating transportation-specific behavioral principles.

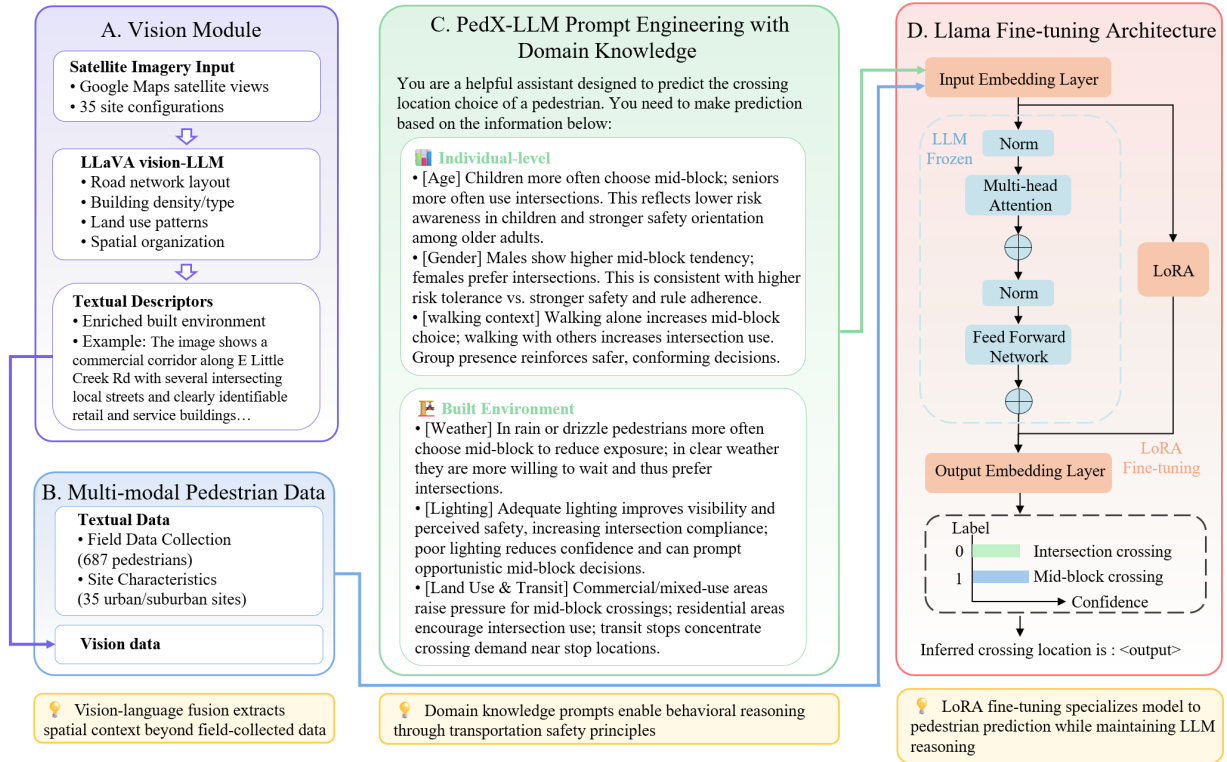

**Figure 3: The PedX-LLM framework architecture.**

## A. Vision Module for Built Environment Extraction

The vision augmentation component (Panel A, Figure 3) leverages the Large Language and Vision Assistant (LLaVA) model (Liu *et al.* 2023) to automatically extract built environment descriptors from satellite imagery. LLaVA integrates three components: (1) a visual encoder using pre-trained CLIP ViT-L/14 (Vision Transformer) that processes 336×336 pixel satellite images, with weights frozen to preserve generalization capabilities acquired through large-scale pre-training; (2) a projection network comprising a two-layer MLP with GELU activation that maps 1024-dimensional visual embeddings into the 4096-dimensional language model embedding space; and (3) a language model (Vicuna-v1.5-7B) that generates natural language descriptions from the projected visual features.

Google Maps satellite imagery was retrieved for all 35 sites at zoom level 19, capturing approximately 50×50 meter areas encompassing the mid-block crossing location and surrounding urban context, for example as illustrated in Figure 2 (b). Each image was processed through LLaVA using structured prompts: "Describe the urban environment visible in this satellite image, focusing on: (1) road network layout and organization, (2) building density and types visible, (3) land use patterns you can identify, (4) spatial organization of the area." The model generates textual descriptors averaging 150-200 words per site, capturing macro-scale spatial characteristics. Example output: "The image shows a commercial corridor along E Little Creek Rd with several intersecting local streets and clearly identifiable retail and service buildings. Parking lots surround many establishments, indicating an auto-oriented layout. The area includes convenience stores, clinics, and small businesses, suggesting steady pedestrian activity within a moderately developed urban setting."

These vision-derived context descriptions provide qualitative spatial information, such as building scale relationships, development patterns, and overall urban form, complementing quantitative textual data. This enables the downstream language model to reason about built environment influences on pedestrian crossing decisions.

## B. Multi-modal Pedestrian Data

The multi-modal pedestrian data component (Panel B, Figure 3) integrates visual data from satellite imagery and textual data from field observations. As detailed in Section 3, the dataset encompasses 687 pedestrian crossing observations spanning 35 urban and suburban sites in Hampton Roads, Virginia. The textual data component captured individual-level attributes including pedestrian demographics, walking context, environmental conditions, and observed crossing location choices as summarized in Table 2. Built environment features were systematically documented to capture roadway geometry, traffic control features, and land use context. Specific variables include lane counts, speed limits, signal timing parameters, and categorical land use definitions, as summarized in Table 1. These field-collected textual data provide ground-truth behavioral observations and quantitative infrastructure measurements. Meanwhile, the visual data from satellite imagery (processed through the LLaVA vision-LLM as described in Section 4.A) extracts spatial context including road network layout, building density patterns, land use configurations, and spatial organization. Synthesizing these textual field metrics with vision-derived spatial descriptors constructs a holistic multi-modal representation of crossing scenarios. This integration addresses limitations of conventional approaches that rely exclusively on either observational data or infrastructure inventories.

## C. PedX-LLM Prompt Engineering with Domain Knowledge

Domain knowledge represents empirically validated prior knowledge from established research. In this framework, domain knowledge refers to behavioral principles regarding pedestrian demographics (age, gender, walking context) and built environment factors (lighting, land use, traffic control) derived from the Virginia DOT Hampton Roads pedestrian crossing study (Xie *et al.* 2024). The collected multi-modal data are transformed into the PedX-LLM Prompt through systematic textualization integrating this domain knowledge with site-specific observations. As illustrated in Panel C of Figure 3, the domain knowledge injection is organized into two hierarchical categories: individual-level attributes and built environment.

The system prompt explicitly encodes behavioral priors derived from these empirical studies. Individual-level attributes integrate pedestrian demographics with validated behavioral patterns. Specifically, the prompt instructs the model that age significantly influences crossing choice, noting that older adults typically demonstrate higher safety awareness and delay tolerance (correlating with a lower probability of mid-block crossing). Regarding gender, the prompt encodes findings that male participants exhibit a higher tendency for mid-block crossing compared to females. Similarly, for walking context, the system is primed to recognize that pedestrians walking alone are statistically more likely to cross at mid-block locations compared to those in groups.

Regarding the built environment, the prompt incorporates documented environmental determinants. Lighting conditions are encoded as a key factor, with the instruction that adequate sight distance and lighting encourage mid-block crossings by improving perceived safety. For land use and transit, the model is guided by the principle that specific interactions (e.g., office-residential) generate lower crossing demand, while pedestrian generators like bus stops concentrate crossing activities, necessitating distinct safety considerations.

To operationalize these principles, the framework organizes multi-modal data into structured text-based inputs for training and inference. As detailed in Table 3, data are integrated from four sources: field observations, built environment, vision-derived spatial context, and domain knowledge. Each prompt component systematically presents attribute information, concluding with an inference task specification that directs the model to generate a binary classification output (0 = intersection crossing, 1 = mid-block crossing). This hierarchical structure enables the language model to synthesize domain knowledge with site-specific contextual information, aligning with principles of transparent AI inference where explicit domain knowledge integration enhances the interpretability of inferred behavioral mechanisms.

**Table 3 Input Data Structure for PedX-LLM Framework**

| Prompt Component | Data Source | Variable Name |
|---|---|---|
| Domain Knowledge Context | (Section 4.C) | |
| Pedestrian Demographics | Table 2 | Age group |
| | Table 2 | Gender |
| | Table 2 | Walking context |
| Environmental Conditions | Table 2 | Weather condition |
| | Table 2 | Lighting at intersection |
| | Table 2 | Lighting at mid-block |
| Roadway Geometry | Table 1 | Number of lanes at mid-block |
| | Table 1 | Speed limit |
| | Table 1 | Total width of sidewalk |
| | Table 1 | Presence of raised median |
| Traffic Control | Table 2 | Time interval between consecutive onsets of green time for crossing |
| | Table 2 | Green interval for crossing |
| | Table 2 | Pedestrian push-button available |
| | Table 2 | Push-button affecting crossing time |
| Land Use & Transit | Table 1 | Land use |
| | Table 1 | Presence of Public Transit Station |
| Built Environment (Vision) | (Section 4.A) | |

## D. LLM Fine-tuning Architecture and Training Configuration

The PedX-LLM framework employs LoRA (Hu *et al.* 2022) to fine-tune the LLaMA-2-7B model. This architecture is selected specifically to address the computational and privacy constraints of transportation agencies. Unlike full fine-tuning, which retrains all model parameters, LoRA introduces a "bypass" mechanism that allows the model to learn new pedestrian crossing behaviors while keeping its vast pre-trained linguistic knowledge intact.

To enable task-specific learning without catastrophic forgetting, we freeze the pre-trained weights ( $W_0$ ) of the LLaMA base model. Trainable low-rank matrices are then injected into the attention layers to represent the domain-specific adaptation.

Mathematically, for an input $x$, the forward pass output $h$ is computed as the sum of the general linguistic knowledge (fixed) and the transportation domain adaptation (trainable):

$$h = W_0 x + \frac{\alpha}{r} BAx \tag{1}$$

$W_0 x$ represents the frozen general knowledge retained from the base model (e.g., understanding English grammar and logic). $BAx$ represents the learned domain knowledge, where $B \in \mathbb{R}^{d \times r}$ and $A \in \mathbb{R}^{r \times d}$ are low-rank matrices (rank $r = 32$ ). $\frac{\alpha}{r}$ is a scaling factor that controls the influence of the new domain knowledge on the final decision.

This decomposition ensures that the model learns the nuances of pedestrian behavior (via matrices $A$ and $B$ ) without overwriting its fundamental reasoning capabilities. This configuration results in approximately 32.5 million trainable parameters, representing only 0.46% of the 7 billion total parameters in LLaMA-2-7B, making it feasible for deployment on standard agency workstations.

To further reduce the memory footprint from 28 GB to approximately 7 GB, we utilize 4-bit NormalFloat (nf4) quantization, which compresses model weights to 4-bit precision while maintaining computational accuracy through mixed-precision calculations. The model is optimized using a causal language modeling objective. The tokenizer vocabulary is expanded with domain-specific tokens (<INTERSECTION>, <MIDBLOCK>), and the loss function minimizes the negative log-likelihood strictly on these answer tokens:

$$L = -\sum_{i=1}^{N} I_{\text{answer}}(i) \log P(t_i \mid t_1, \ldots, t_{i-1}; \theta) \tag{2}$$

By masking the prompt context ( $I_{\text{answer}} = 0$ ), we ensure the model learns exclusively from answer tokens, forcing it to develop generalizable reasoning patterns rather than memorizing prompt structures. Detailed hyperparameters, including the AdamW optimizer settings and the cosine learning rate schedule, are summarized in Table 4.

**Table 4 Model architecture and training configuration**

| Category | Parameter | Configuration / Value |
|---|---|---|
| Base Model | Architecture | LLaMA-2-7B-hf |
| | Hidden Size ( $d$ ) | 4096 |
| Quantization | Precision Type | 4-bit NormalFloat (nf4) |
| | Computation Dtype | float16 |
| LoRA Adapter | Rank ( $r$ ) | 32 |
| | Alpha ( $\alpha$ ) | 64 |
| | Scaling Factor | 2.0 |
| | Target Modules | q_proj,k_proj, v_proj, o_proj |
| | Trainable Parameters | $\approx$32.5 Million (0.46%) |
| | Dropout Rate | 0.1 |
| Optimization | Optimizer | AdamW |
| | Learning Rate | $2 \times 10^{-5}$ |
| | Weight Decay | 0.01 |
| | LR Schedule | Cosine with 50-step Warmup |
| | Gradient Clipping | 0.5 |
| Training | Effective Batch Size | 16 |
| | Max Sequence Length | 512 tokens |
| | Max Epochs | 250 |
| | Early Stopping Patience | 15 epochs |
| | Evaluation Metric | Balanced Accuracy |
| Hardware | GPU | NVIDIA Quadro RTX 5000 ( 32 GB) |

## 5. Results and Discussion

### 5.1. Model Performance and Baseline Comparisons

To rigorously evaluate the efficacy of the proposed framework, comprehensive comparisons were conducted against a spectrum of established baseline methods representing distinct methodological paradigms. These baselines are categorized into two tiers: (1) Statistical Models, including Logistic Regression, Hierarchical Logistic Regression serving as the standard for behavioral interpretability; and (2) Supervised Learning Baselines, comprising Random Forest, XGBoost, Support Vector Machine (SVM), CatBoost, selected to represent widely adopted benchmarks for predictive accuracy on tabular data, Multi-layer Perceptron (MLP), and TabNet, utilized to assess the capacity of advanced architectures in handling high-dimensional feature interactions and textual inputs. The dataset of 687 observations was partitioned into training (70%), validation (15%), and test (15%) sets via stratified random sampling to strictly maintain the inherent class imbalance (37.7% mid-block vs. 62.3% intersection) across all partitions. All models were trained and hyperparameter-tuned using the validation set, with final performance metrics (mean and standard deviation) computed on the held-out test set across five random splits.

Table 5 presents the performance metrics computed across five random splits. Given the class imbalance (37.7% mid-block vs. 62.3% intersection), this study reports balanced accuracy as the primary metric, defined as the arithmetic mean of sensitivity and specificity, ensuring equal weight to both classes. Additional metrics include: Accuracy (overall correct predictions), Precision (TP/(TP+FP), where mid-block crossing is the positive class), F1-Score (harmonic mean of precision and recall). Standard Deviation (Std Dev) quantifies the variability of balanced accuracy across the five random splits, indicating model stability.

**Table 5 Comparative performance of pedestrian crossing location inference models**

| Model | Accuracy | Precision | F1-Score | Balanced Accuracy | Std Dev |
|---|---|---|---|---|---|
| **Statistical Models** | | | | | |
| Logistic Regression | 73.5% | 70.2% | 72.6% | 73.2% | 2.3% |
| Hierarchical Logistic Regression | 74.2% | 71.5% | 72.7% | 74.1% | 2.2% |
| **Supervised Learning** | | | | | |
| Random Forest | 77.3% | 74.9% | 77.0% | 77.1% | 2.1% |
| XGBoost | 78.9% | 76.4% | 78.3% | 78.7% | 1.9% |
| Support Vector Machine | 75.1% | 72.6% | 74.7% | 75.0% | 2.7% |
| CatBoost | 79.2% | 76.9% | 78.8% | 79.0% | 2.0% |
| Multi-layer Perceptron | 78.0% | 75.3% | 77.6% | 77.8% | 2.2% |
| TabNet | 79.7% | 77.2% | 79.2% | 79.4% | 1.9% |
| **LLM Frameworks** | | | | | |
| Baseline LLaMA-2-7B | 62.3% | 59.9% | 62.0% | 62.1% | 3.1% |
| PedX-LLM (Text-only) | 75.2% | 72.7% | 74.8% | 75.0% | 1.8% |
| PedX-LLM (Vision-Augmented) | 77.4% | 74.2% | 76.7% | 77.9% | 1.7% |
| **PedX-LLM (Vision-and-Knowledge Augmented)** | **82.1%** | **80.0%** | **81.7%** | **82.0%** | **1.4%** |

Statistical models and supervised learning baselines establish a competitive performance standard. Statistical approaches achieved balanced accuracy scores ranging from 73.2% to 74.1%, with Hierarchical Logistic Regression performing best. Supervised learning baselines showed improved performance across all metrics. CatBoost achieved the highest balanced accuracy of 79.0%, with precision of 76.9%, F1-score of 78.8%. TabNet achieved 79.4% balanced accuracy with 79.7% overall accuracy and 79.2% F1-score.

The Baseline LLaMA-2-7B performs poorly with 62.1% accuracy, reflecting a significant domain gap between the model's pre-training data (general web text) and our specialized transportation safety context. Supervised fine-tuning via LoRA significantly narrows this gap. The PedX-LLM (Text-only) variant improved balanced accuracy to 75.0% (12.9 percentage points gain), with corresponding improvements in precision to 72.7% and F1-score to 74.8%. However, this text-only approach still failed to surpass strong baselines like CatBoost (79.0%), suggesting that pure pattern recognition is insufficient for this task.

The integration of vision-derived built environment context further optimizes performance. The PedX-LLM (Vision-Augmented) variant utilizes satellite imagery features without domain knowledge and achieved 77.9% balanced accuracy, representing a 2.9% gain over the text-only version. These results confirm that automatically extracted visual features capture macro-scale spatial contexts that complement manual site data.

Building upon PedX-LLM (Vision-Augmented), the PedX-LLM (Vision-and-Knowledge Augmented) incorporates transportation domain knowledge to achieve a peak balanced accuracy of 82.0%, representing a 4.1 percentage point improvement over the Vision-Augmented configuration. This demonstrates the critical role of domain knowledge in enhancing behavioral reasoning beyond multi-modal data fusion alone. The full framework surpasses the best statistical model (Hierarchical Logistic Regression) by 7.9 percentage points and the strongest Supervised learning baseline (CatBoost) by 3.0 percentage points, while maintaining the lowest standard deviation of 1.4%, indicating robust generalization across diverse site configurations.

## 5.2. Ablation Study of Domain Knowledge

To quantify the contribution of domain knowledge prompts (Panel C, Figure 3), This section conducted systematic ablation experiments on two knowledge categories: Individual-level knowledge (age, gender, and walking context) and Built Environment knowledge (weather, lighting, and land use & transit). By selectively integrating these components individually and in combination, we isolated their respective contributions to crossing location inference performance. Table 6 presents the ablation results, where Category (Δ) represents each knowledge category's improvement over the baseline and Factor Contribution (Δ) quantifies each individual factor's contribution (calculated as its within-category weight multiplied by the category's overall gain), demonstrating that the fully integrated framework achieves the highest accuracy.

**Table 6 Ablation analysis of domain knowledge contributions**

| Knowledge Category | Knowledge Factors | Factor Contribution (Δ) | Balanced Accuracy | Category (Δ) |
|---|---|---|---|---|
| PedX-LLM (Vision-Augmented) | None | | 77.9% | |
| +Individual-level | Age | +1.19% | 80.9% | +3.0% |
| | Gender | +0.80% | | |
| | Walking context | +1.01% | | |
| +Built Environment | Weather | +1.27% | 81.2% | +3.3% |
| | Lighting | +1.11% | | |
| | Land Use & Transit | +0.92% | | |
| PedX-LLM (Vision-and-Knowledge Augmented) | Both | | 82.0% | +4.1% |

Note: Improvement (Δ) is calculated relative to the baseline (PedX-LLM (Vision-Augmented)).

The baseline PedX-LLM (Vision-Augmented) model, utilizing only visual and textual features without domain knowledge prompts, established a balanced accuracy of 77.9%. The ablation results reveal distinct contributions from the two knowledge categories. Integrating Individual-level knowledge (age, gender, and walking context) provided a moderate performance boost of 3.0 percentage points, elevating accuracy to 80.9%. In comparison, incorporating Built Environment knowledge (weather, lighting, and land use & transit) yielded a stronger individual gain of 3.3 percentage points, achieving 81.2% accuracy. This suggests that environmental context contributes more decisively to behavioral inference than individual demographics when each category is considered in isolation. The complete PedX-LLM (Vision-and-Knowledge Augmented) framework, synthesizing both knowledge streams, achieves the highest accuracy of 82.0% with a combined gain of 4.1 percentage points, demonstrating that integrating both knowledge categories substantially enhances inference performance.

To understand the specific drivers behind these category-level gains, this section decomposed each knowledge category into its constituent factors using permutation-based feature attribution. Within the Individual-level knowledge category, Age emerges as the dominant factor, contributing 1.19 percentage points (39.6% of the category's total gain). This reflects the critical variance in risk perception and crossing capability across age groups, particularly between children and seniors who exhibit markedly different safety orientations. Walking context follows as the second most influential attribute with a 1.01 percentage point contribution (33.5%), capturing the safety benefits of social conformity and collective decision-making when pedestrians walk in groups. Gender accounts for 0.80 percentage points (26.9%), reflecting documented differences in risk tolerance between male and female pedestrians.

The Built Environment knowledge category reveals a different hierarchy of importance. Weather is identified as the primary environmental driver, contributing 1.27 percentage points (38.7% of the category's gain). This dominance is consistent with behavioral principles where adverse weather conditions elevate the perceived cost of waiting at intersections, thereby increasing the utility of mid-block crossing as a time-saving strategy. Lighting contributes 1.11 percentage points (33.5%), confirming that visibility constraints fundamentally shape pedestrians' crossing location decisions by affecting both perceived safety and the detectability of crossing opportunities. Land Use & Transit adds 0.92 percentage points (27.8%), reflecting how the proximity of transit stops and commercial activities concentrate crossing demand and influence route choices. These results validate that integrating both knowledge categories constructs a robust framework for inferring pedestrian crossing behavior, substantially outperforming models that rely on either knowledge stream alone.

### 5.3. Evaluate Cross-Site Generalizability

Generalizing pedestrian behavior models to unseen locations with varying geometric and operational characteristics presents a significant challenge. Models trained on specific sites frequently overfit on local features, resulting in poor cross-site generalization. A systematic cross-site validation utilizing a site-based partitioning strategy evaluated the robustness of the PedX-LLM Framework, ensuring complete separation between training and testing environments.

Three non-overlapping groups were created from the dataset of 687 observations across 35 sites in Hampton Roads to prevent data leakage. The training set comprised 22 sites (N=455), covering a diverse range of urban arterials and suburban collectors. The validation set included 5 sites (N=102) for hyperparameter tuning. The test set consisted of 5 distinct sites (N=130) selected to represent the most challenging scenarios, characterized by significant geometric variation.

Benchmarking against established baselines provided context for the framework's performance. Beyond traditional algorithms (Logistic Regression, Hierarchical Logistic Regression, CatBoost, and TabNet), the generalization capabilities of the proposed framework were evaluated using two distinct inference configurations. The PedX-LLM (Zero-shot) configuration applies domain-knowledge-enhanced prompts directly to unseen test sites without site-specific examples. Alternatively, the PedX-LLM (Few-shot) configuration incorporates five randomly selected examples from the validation dataset into the prompt context, facilitating adaptation via in-context learning (ICL) without further parameter updates. Table 7 presents the aggregate performance metrics on the held-out test set.

**Table 7 Cross-site validation results on held-out test set (5 sites, 130 observations)**

| Model | Accuracy | Precision | F1-Score | Balanced Accuracy | Std Dev |
|---|---|---|---|---|---|
| Logistic Regression | 58.5% | 59.6% | 48.7% | 41.2% | 4.8% |
| Hierarchical Logistic Regression | 61.4% | 60.1% | 60.6% | 46.2% | 4.5% |
| CatBoost | 62.1% | 64.5% | 55.3% | 48.3% | 4.2% |
| TabNet | 69.7% | 61.9% | 51.1% | 43.6% | 4.6% |
| PedX-LLM (Zero-shot) | 75.8% | 68.5% | 67.6% | 66.9% | 4.1% |
| PedX-LLM (Few-shot) | 79.4% | 73.0% | 72.6% | 72.2% | 3.5% |

The PedX-LLM Framework (Zero-shot) achieved a balanced accuracy of 66.9%, outperforming the baseline (CatBoost, 48.3%) by 18.6 percentage points. Traditional supervised learning models exhibited systematic bias toward the majority class; notably, TabNet achieved relatively high overall accuracy (69.7%) but a low balanced accuracy (43.6%). Conversely, PedX-LLM maintained superior balance across all metrics, with F1-scores substantially exceeding baselines. Incorporating five validation examples via the Few-shot configuration further improved balanced accuracy to 72.2% (+5.3%) and reduced performance variance (Std Dev: 3.5%), demonstrating robust adaptability with minimal calibration data.

To enable a detailed assessment of cross-site generalization performance, Table 8 presents model performance on five unseen test sites ranging from 2-lane to 10-lane facilities, encompassing both simple and complex crossing scenarios.

**Table 8 Model generalization performance on unseen test sites**

| Site ID | Location | Lanes | Logistic Regression | Hierarchical Logistic Regression | CatBoost | TabNet | PedX-LLM (Zero-shot) | PedX-LLM (Few-shot) |
|---|---|---|---|---|---|---|---|---|
| 1 | Hampton Blvd | 6 | 38.9% | 42.5% | 46.1% | 41.6% | 65.3% | 69.1% |
| 9 | Azalea Garden Rd | 2 | 45.1% | 48.0% | 51.4% | 46.5% | 70.7% | 74.7% |
| 10 | N Military Hwy | 10 | 35.2% | 39.1% | 44.6% | 36.9% | 61.8% | 67.7% |
| 27 | Hampton Hwy | 4 | 38.5% | 41.7% | 45.2% | 40.0% | 62.6% | 68.8% |
| 35 | Bridge Rd | 4 | 47.7% | 51.1% | 54.0% | 49.3% | 73.8% | 76.5% |
| Mean | | | 41.2% | 44.8% | 48.3% | 43.6% | 66.9% | 72.2% |
| Std Dev | | | 4.8% | 4.7% | 3.8% | 4.9% | 5.0% | 3.8% |

Consistent superiority of the PedX-LLM Framework is evident across all test environments, yielding Zero-shot performance gains ranging from 17.2% at Site 10 to 19.8% at Site 35 compared to the supervised learning baseline, CatBoost. Site 10, a 10-lane arterial with a low mid-block crossing rate (12%), represented the most challenging scenario due to its extreme geometry. Baseline models failed to generalize, yielding balanced accuracy scores between 35.2% and 44.6%, whereas PedX-LLM achieved 61.8% in the Zero-shot configuration. This success stems from the model's ability to apply domain knowledge regarding the deterrent effect of excessive lane counts on crossing behavior. Similarly, the model effectively adapted to distribution shifts at Site 9 (a 2-lane collector with the highest mid-block rate), achieving 70.7% accuracy compared to 51.4% for CatBoost.

Baseline model performance varied substantially across sites (35.2%-54.0% balanced accuracy), with poorest performance at extreme geometric configurations. Site 10 (10-lane arterial) exhibited the lowest baseline accuracy (35.2%-44.6%) due to insufficient training data coverage of rare wide-roadway configurations, whereas Site 35 (4-lane collector) showed relatively better baseline performance (47.7%-54.0%) as this geometry more closely matched typical training conditions. PedX-LLM maintained consistent performance across all configurations through domain-knowledge-enhanced reasoning.

These results confirm that the domain-knowledge-enhanced linguistic reasoning of the PedX-LLM Framework enables the model to generalize reliably across diverse sites. Transforming site-specific pattern recognition into generalizable behavioral inference overcomes the distribution shift limitations inherent in purely data-driven approaches, offering a scalable solution for diverse urban environments.

### 5.4. Interpretability and Demonstration

To interpret how the PedX-LLM Framework infers pedestrian crossing behavior, this section employed a sentence-based feature attribution method grounded in Shapley value theory (Winter 2002). Originating from cooperative game theory, Shapley values provide a mathematically rigorous framework for attributing a model's output to its input features by calculating each feature's average marginal contribution across all possible feature combinations. This approach ensures fair attribution through key properties including efficiency (attributions sum to the model output) and symmetry (equivalent features receive equal credit). Widely adopted for supervised learning interpretability, Shapley values have proven particularly effective for understanding complex model behaviors. In our implementation, the input prompt is decomposed into seven distinct components (Pedestrian Demographics, Traffic Control, Roadway Geometry, Built Environment, Land Use & Transit, Environmental Conditions, and Domain Knowledge Context). Each component is treated as a feature, and its Shapley value quantifies the component's contribution to the model's crossing location inference confidence.

This study quantified global feature importance by aggregating the absolute Shapley values across the entire dataset of 687 observations. Table 9 presents the ranked contributions of each prompt component.

**Table 9 Aggregate feature importance in crossing location inference**

| Rank | Prompt Component | Average Absolute Shapley Value | Contribution (%) |
|---|---|---|---|
| 1 | Pedestrian Demographics | 0.304 | 25.8% |
| 2 | Traffic Control | 0.257 | 21.8% |
| 3 | Domain Knowledge Context | 0.150 | 12.7% |
| 4 | Roadway Geometry | 0.147 | 12.5% |
| 5 | Built Environment (Vision) | 0.142 | 12.1% |
| 6 | Land Use & Transit | 0.090 | 7.6% |
| 7 | Environmental Conditions | 0.088 | 7.5% |

Pedestrian Demographics (25.8%) and Traffic Control (21.8%) are the dominant factors, collectively accounting for nearly half of the model's decision-making weight. This aligns with behavioral literature suggesting that individual risk tolerance and delay are primary motivators. Domain Knowledge Context (12.7%), Roadway Geometry (12.5%), and Built Environment (12.1%) form a strong secondary tier of determinants. This highlights that while individual traits drive decisions, the model relies heavily on the physical framework and domain principles to modulate these choices. Land Use & Transit (7.6%) and Environmental Conditions (7.5%) provide situational context that further refines the inference.

To demonstrate the model's ability to capture critical demographic and built environment differences in predicting crossing behavior, Figure 4 visualizes feature attribution results for representative cases. Each case comprises four components: the Original Data section (left), including Textual Data and Vision Data; the PedX-LLM Input section displays the structured prompt fed into the model; the Model Inference Process section (middle) explains the model's reasoning for how different factors influence the crossing decision; and the Inferred Results box with Shapley Values chart (right) displays the predicted crossing location, model confidence, and feature contributions where negative values (left side) favor intersection crossing and positive values (right side) favor mid-block crossing.

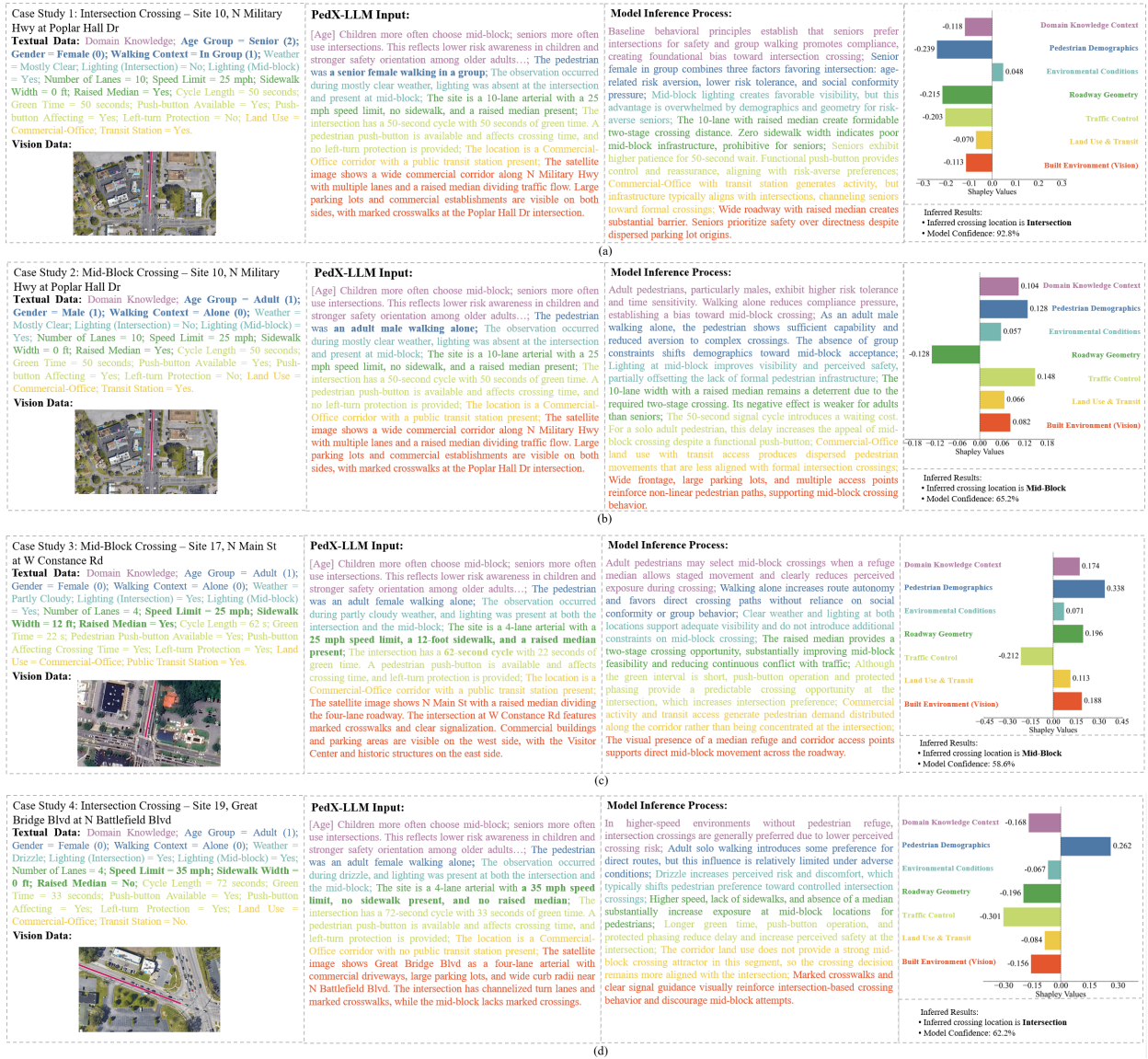

**Figure 4: Feature attribution for individual cases: (a–b) Same built environments with different pedestrian demographics (senior female in a group vs. adult male alone); (c–d) same pedestrian demographics with different built environments (Sites 17 and 19).**

Cases 1 and 2 demonstrate how identical built environment conditions at Site 10 produce opposite crossing predictions (92.8% intersection versus 65.2% mid-block) driven entirely by pedestrian demographic differences. Pedestrian Demographics exhibits the most dramatic reversal (−0.239 to +0.128), reflecting shifts from senior female group caution to adult male solo risk tolerance. Domain Knowledge Context reverses direction (−0.118 to +0.104), adapting from senior intersection preference to adult mid-block acceptance. Traffic Control maintains identical 50-second cycles yet produces opposite effects (−0.203 versus +0.148), where seniors perceive waiting as acceptable while solo adults view delays as mid-block incentives. Roadway Geometry remains consistently negative (−0.215 versus −0.128), with the 10-lane arterial creating prohibitive barriers for seniors but only moderate deterrents for adults. Environmental Conditions demonstrates context-dependent interpretation (+0.048 versus +0.057), where mid-block lighting advantages are overwhelmed for risk-averse seniors but actively support adult decisions. Land Use & Transit and Built Environment reverse from negative to positive (−0.070 to +0.066; −0.113 to +0.082), channeling seniors toward formal crossings while dispersed commercial layouts reinforce adult mid-block paths. This validates that the model modulates identical infrastructure effects based on demographic vulnerability rather than applying fixed weights.

Cases 3 and 4 demonstrate how different built environments produce opposite crossing predictions (58.6% mid-block versus 62.2% intersection) despite identical pedestrian demographics (adult female walking alone). Pedestrian Demographics maintains consistent positive contributions (+0.338 versus +0.262), reflecting adult solo preference for direct routes. Traffic Control exhibits the most substantial reversal from mid-block facilitator to strongest intersection predictor (−0.212 to −0.301), where Site 17's short green intervals increase intersection preference despite predictable phasing, while Site 19's longer green time and protected phasing reduce delays and enhance perceived intersection safety. Roadway Geometry reverses from facilitator to deterrent (+0.196 to −0.196), with Site 17's raised median enabling safe two-stage crossings while Site 19's median absence combined with higher speed and zero sidewalk width substantially increases mid-block exposure. Environmental Conditions shifts from positive to negative (+0.071 to −0.067), as Site 17's clear weather supports adequate visibility without additional constraints, whereas Site 19's drizzle increases perceived risk shifting preference toward controlled intersections. Built Environment reverses direction (+0.188 to −0.156), where Site 17's median refuge and corridor access points support direct mid-block movement while Site 19's marked crosswalks and signal guidance reinforce intersection behavior. Domain Knowledge Context and Land Use & Transit similarly reverse (+0.174 to −0.168; +0.113 to −0.084), with Site 17's commercial activity and transit access distributing pedestrian demand along corridors versus Site 19's higher-speed environment lacking strong mid-block attractors. This validates that roadway geometry emerges as the critical modulator, where infrastructure deficits amplify adverse environmental conditions.

Integrating the interpretability analysis with case-level attributions reveals actionable intervention priorities for sites exhibiting high mid-block crossing rates: (1) optimizing signal timing to reduce pedestrian delay at adjacent intersections, thereby decreasing the utility gap between crossing locations, as Traffic Control consistently ranks among the top contributors; (2) installing physical deterrents such as median barriers or fencing to increase perceived mid-block crossing complexity, given the substantial influence of Roadway Geometry demonstrated across cases; (3) enhancing intersection infrastructure with responsive push-buttons and clearly marked crosswalks to improve compliant crossing attractiveness, particularly for vulnerable populations. The model's capacity to jointly consider geometric features, traffic control parameters, and demographic characteristics confirms that vision-and-knowledge-enhanced reasoning enables interpretable, context-aware behavioral inference, providing a foundation for targeted, site-specific safety interventions.

## 6. Conclusion

This study introduces PedX-LLM, a framework designed to shift pedestrian crossing inference from numeric pattern fitting to generalizable context-aware behavioral reasoning. The framework integrates multimodal inputs, including satellite imagery (to capture the built environment) and textual data, and encodes domain knowledge of pedestrian behavior to construct a comprehensive representation of pedestrian crossing decisions. We employ LoRA to fine-tune the framework using training data and evaluate its performance in previously unseen environments.

The findings demonstrate the effectiveness of the PedX-LLM framework in inferring pedestrian crossing behavior and identifying key determinants. The fully integrated framework, combining individual-level attributes, built environment features, and domain knowledge, achieved a balanced accuracy of 82.0%, exceeding CatBoost by 3.0 percentage points and the statistical model (Hierarchical Logistic Regression) by 7.9 percentage points. Shapley-based attribution analysis identifies Pedestrian Demographics (25.8%) and Traffic Control (21.8%) as the primary behavioral drivers. Incorporating vision-augmented built environment features extracted from satellite imagery through the LLaVA module contributed an additional 2.9% performance gain, demonstrating the value of visual context integration. Ablation studies quantified the contribution of domain knowledge integration; combining individual-level and environmental knowledge produced a cumulative improvement of 4.1 percentage points over the baseline. Moreover, cross-site evaluation confirmed the model's robust generalization capability. In zero-shot configurations on unseen sites, the framework achieved 66.9% balanced accuracy, significantly outperforming Hierarchical Logistic Regression (46.2%) and CatBoost (48.3%). Few-shot adaptation further elevated performance to 72.2%, demonstrating that the model effectively overcomes the distribution shifts that constrain purely data-driven approaches.

This study contributes to the literature by establishing a paradigm for specializing LLMs in pedestrian behavior analysis. First, it introduces a multimodal architecture that integrates vision-derived contextual information with textual behavioral records, overcoming the limitations of unimodal approaches that fail to capture synergistic environmental effects. Second, by employing LoRA for parameter-efficient fine-tuning, the proposed framework enables effective utilization of local data to achieve improved performance. Third, the framework embeds transportation domain knowledge into the model's reasoning process via structured prompt engineering. This integration elevates site-specific pattern recognition to generalizable behavioral inference. Finally, the framework is implemented using open-source LLMs deployed on local servers, ensuring protection of proprietary data and preservation of privacy.

From a practical perspective, PedX-LLM supports pedestrian behavior analysis in data-scarce and heterogeneous urban environments where traditional models perform poorly. Its strong zero-shot and few-shot generalizability enables transportation engineers to assess pedestrian crossing behavior at new or modified sites with minimal local data. By jointly leveraging visual context and domain knowledge, the framework also facilitates design evaluation and scenario-based analysis of pedestrian facilities, informing infrastructure and traffic control decisions. Moreover, deployment on locally hosted open-source LLMs provides a privacy-preserving and extensible foundation for agency-facing decision-support tools.

Future work will focus on expanding the dataset to encompass more diverse urban contexts, including high-density metropolitan areas and suburban corridors, validating the framework across different geographic regions with varying traffic regulations and cultural norms, and exploring the integration of real-time data streams such as traffic volumes and weather conditions to enable dynamic behavioral inferences and adaptive countermeasure deployment.

## Acknowledgment

The work was partially funded by the Transportation Informatics Lab, Department of Civil & Environmental Engineering at Old Dominion University (ODU). The pedestrian crossing data used in this study were collected as part of the project *"Factors Influencing Pedestrian Decisions to Cross Mid-Block and Potential Countermeasures,"* funded by the Virginia Transportation Research Council (VTRC). The contents of this paper present the views of the authors, who are responsible for the facts and accuracy of the data presented herein. The contents of the paper do not reflect the official views or policies of VTRC or other associated agencies.